\pdfoutput=1
\documentclass[11pt]{article}

\usepackage{acl}

\usepackage{times}
\usepackage{latexsym}
\usepackage{amsfonts}
\usepackage[T1]{fontenc}
\usepackage[utf8]{inputenc}

\usepackage{microtype}

\usepackage{graphicx}
\usepackage{float}
\usepackage{hhline}
\usepackage{multirow}
\usepackage{multicol}
\usepackage{amsmath}
\usepackage{hyperref}
\makeatletter
\newcommand{\printfnsymbol}[1]{%
  \textsuperscript{\@fnsymbol{#1}}%
}
\makeatother

\title{AdapterBias: Parameter-efficient Token-dependent Representation Shift for Adapters in NLP Tasks}

\author{Chin-Lun Fu$^{1\star}$\qquad  Zih-Ching Chen$^{2\star}$\qquad  Yun-Ru Lee$^{3}$\qquad  Hung-yi Lee$^{2}$ \\
$^{1}$Department of Electrical Engineering, National Taiwan University\\
$^{2}$Graduate Institute of Communication Engineering, National Taiwan University\\
$^{3}$Department of Computer Science and Information Engineering, National Taiwan University\\
         \texttt{\{b06505011,r09942176,b06902107,hungyilee\}@ntu.edu.tw}}

\begin{document}

\maketitle
\begin{abstract}
Transformer-based pre-trained models with millions of parameters require large storage. Recent approaches tackle this shortcoming by training adapters, but these approaches still require a relatively large number of parameters. 
 In this study, AdapterBias, a surprisingly simple yet effective adapter architecture, is proposed.
AdapterBias adds a token-dependent shift to the hidden output of transformer layers to adapt to downstream tasks with only a vector and a linear layer.
 Extensive experiments are conducted to demonstrate the effectiveness of AdapterBias. 
 The experiments show that our proposed method can dramatically reduce the trainable parameters compared to the previous works with a minimal decrease in task performances compared with fine-tuned pre-trained models.
We further find that AdapterBias automatically learns to assign more significant representation shifts to the tokens related to the task in consideration.\footnote{The source code is available at: \url{https://github.com/Allen0307/AdapterBias}}

% Transformer-based pre-trained models with millions of parameters require large storage. Recent approaches tackle this shortcoming by training adapters, but these approaches still require a relatively large number of parameters. 
%Many parameter-efficient methods add a shift to the token embedding to adapt to different downstream tasks achieved comparable performance while dramatically decreasing the trained parameter. However, the shifts added are the same for all input tokens in the downstream task. 
 % However, whether these adapters are already at their simplest state remains a question.
 %In this study, AdapterBias, a surprisingly simple yet effective adapter architecture that adds a specific shift based on the input token, is proposed, which includes only a vector and a linear layer. 
 %AdapterBias adds a token-dependent shift to the representation to adapt to downstream tasks.
 %Extensive experiments are conducted to demonstrate the effectiveness of AdapterBias. 
 %The experiments show that our proposed method can dramatically reduce the trainable parameters than the previous works with a minimal decrease in task performances compared with fine-tuned pre-trained models. 
 %We further explore the tokens that AdapterBias considers as more important to specific tasks by visualization.
 %Due to the simplicity of our AdapterBias, we further explain which kinds of tokens are more important to specific tasks.
\end{abstract}

\section{Introduction}

While large pre-trained language models (PLMs) reached state-of-the-art results on natural language processing (NLP) tasks, PLMs require updating all parameters and storing the fully fine-tuned model for each downstream task. These requirements have led to difficulties in real-world applications. Moreover, fine-tuning PLMs on low-resource datasets is subject to instabilities. 

To tackle these shortcomings, Adapters~\citep{houlsby2019parameter}, a more parameter-efficient alternative training strategy for the transformer architecture \citep{vaswani2017attention} have been proposed. Instead of full fine-tuning the whole model, Adapters introduce extra tunable weights and freeze the original parameters of PLM. Adapters demonstrated comparable performance with fully fine-tuning the entire model. Although Adapters solve the problem of the PLM's massive parameters, researchers are curious about how many more parameters are required to reach state-of-the-art performance on standard NLP tasks. The results in \citet{houlsby2019parameter} have shown that the performance on GLUE benchmark \citep{wang2018glue} is almost the same when removing the Adapters in the lower layers, which indicates that not every adapter is useful. It raises the question of whether adapters can be even more parameter-efficient.

\begin{figure}
%\vspace{-20pt}
      \centering
      \includegraphics[width=0.48\textwidth]{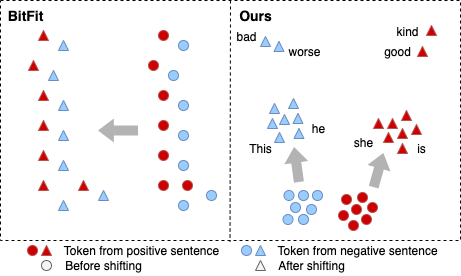}
      \caption{
        Overview of the main concept of our work compared to BitFit \citep{ben2021bitfit}.
        Left: BitFit tends to add the same representation shift to different tokens. 
        Right: Our work applies different representation shifts to tokens considering their importance to the downstream task and their characteristics. 
        The shifts of the input words that are more task-related is more significant than that of other tokens. 
        For example, in SST-2 \citep{socher2013recursive}, which is a semantic task, the representation shifts of the semantic words, such as "kind" and "worse", are larger than that of other words.
      }
\label{fg:overview}
\end{figure}

To develop practical and memory-efficient methods of utilizing PLMs, Diff pruning \cite{guo2020parameter} enables parameter-efficient transfer learning that scales well with new tasks. The approach learns a task-specific “diff” vector that extends the original pre-trained parameters and encourages the sparsity of the vector through $L_0$-norm regularization. Another approach is BitFit \cite{ben2021bitfit}, which shows that with small-to-medium training data, fine-tuning only a subset of the bias terms of pre-trained BERT models \citep{devlin2018bert} is competitive with fine-tuning the entire model. The central concept of these approaches is to add task-specific shifts to each output representation of the PLM layers so as to adapt to different tasks. In the previous works, \citet{ben2021bitfit,guo2020parameter} both add the same shifts to the output representation regardless of which token is more relevant to the task. However, considering some specific tokens might be more critical to a particular task, the representation can better adapt to the downstream task under a limited amount of parameters if these shifts are based on the input tokens.
                
%Specifically, we aim to develop practical, memory-efficient methods that train a minimum set of parameters while achieving performance on par with full fine-tuning for state-of-the-art NLP models this is weird

Based on this concept, in this study, we add token-dependent biases to the shifts by proposing AdapterBias, which consists of a vector and a linear layer ($L_\alpha$). 
The vector represents the task-specific shift, and $L_\alpha$ produces the weights for input tokens. Thus, with the vector and the weights, AdapterBias can add a token-dependent shift to the transformer layer. Since the concept of BitFit \citep{ben2021bitfit} is similar to AdapterBias by adding a shift to the representation, we demonstrate the difference between BitFit and AdapterBias in Figure \ref{fg:overview}. BitFit assigns identical shifts to all the tokens, while AdapterBias adds more significant shifts to the representations that are related to the task.

With fewer trainable parameters required, AdapterBias achieves comparable performance on the GLUE benchmark with \citet{houlsby2019parameter,pfeiffer2020adapterfusion,guo2020parameter,ben2021bitfit, hu2021lora}. We further decrease the parameters of AdapterBias in different ways, including partial weight-sharing in AdapterBias and adding $L_0$-norm regularization. Finally, AdapterBias has better interpretability due to its simplicity. We use different tools, including word cloud and PCA \citep{jolliffe2002springer}, to visualize what AdapterBias has learned, and we found that the proposed approach automatically learns to assign larger representation shifts to the task-related tokens.

%and the linear layer produces the weights of the bias for different tokens,

\section{Related Work}
%\subsection{Adapters}

%[Briefly introduce what adapters are]
%While fine-tuning all the weight from the pretrained model reached the state-of-the-art in many NLP tasks
% 如果有時間的話可以比較一下previous studies的limitation
For NLP tasks, adapters are introduced for the transformer architecture. A set of adapter parameters was added at each transformer layer, which is mostly bottleneck architectures \citet{houlsby2019parameter}. By keeping the output dimension identical, they cause no change to the structure or parameters of the original model. 

Adapters quickly gained popularity in NLP with various applications. For multi-task learning \citep{caruana1997multitask,zhang2017survey,liu2019multi}, a projected self-attention layer is proposed by \citet{stickland2019bert}, while \citet{bapna2019simple} proposed an additional layer norm suitable for machine translation.

Besides the applications of adapters, researchers are also dedicated to improving their performance. Based on the architecture introduced by \citet{houlsby2019parameter}, AdapterFusion \citep{pfeiffer2020adapterfusion} leveraged knowledge from multiple tasks with a new two-stage learning algorithm. Despite the recent popularity of these methods, they still train a relatively large number of training parameters.

%strong performance on GLUE benchmark, 
%An adapter with a two-layer feed-forward bottleneck architecture working well with the pre-trained transformer model is introduced by \citet{houlsby2019parameter}; a \citet{pfeiffer2020adapterfusion} proposed AdapterFusion that leveraged knowledge from multiple tasks with a new two-stage learning algorithm. Despite their recent popularity, there is still room for improvements when it comes to the parameter-efficiency of adapters. 
%Following their previous work  \citet{pfeiffer2020adapterfusion}, they also explore beyond parameter-efficient introduced AdapterDrop \cite{ruckle2020adapterdrop},

Recently, studies start to focus on improving the parameter-efficiency of adaptation to a new task~\citep{yang2021voice2series}. Diff-pruning \citep{guo2020parameter} achieves parameter efficiency by adding a sparse, task-specific difference-vector to the fixed original parameters. The vector is adaptively pruned during training with a differentiable approximation to the $L_0$-norm penalty to encourage sparsity. \citet{ruckle2020adapterdrop} introduced AdapterDrop, which has been recently integrated into AdapterHub \citep{pfeiffer2020adapterhub}. It removes adapters from lower transformer layers during training and inference, which can dynamically reduce the computational cost. \citet{mahabadi2021compacter} proposed Compacter, which improved the trade-off between performance and trainable parameters per task with low-rank optimization.

On the other hand, without modifying the architecture of the PLM, BitFit \citep{ben2021bitfit} shows that fine-tuning only the bias terms of a large PLM is also competitive with fine-tuning the entire model. 
Fine-tuning only the bias terms can be considered as adding a task-specific shift to the token representation.
BitFit is most similar to our work. 
While in BitFit, the shifts added to all the representations are exactly the same for all input tokens, in our work, the shifts are token-dependent.

\begin{figure*}

        \centering
      \includegraphics[width=0.9\textwidth]{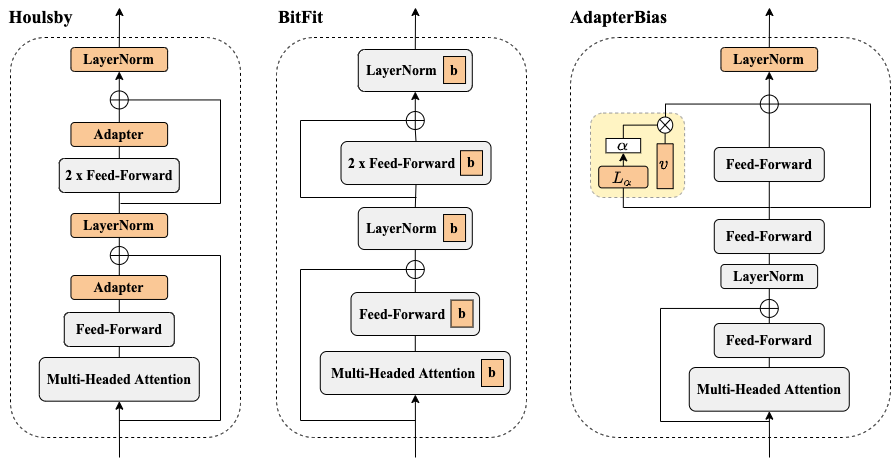}
      
      \caption{Model architectures comparison of \citet{houlsby2019parameter}, BitFit \citep{ben2021bitfit}, and the proposed method AdapterBias. The orange blocks indicate the trainable parts, while the gray blocks indicate the frozen parameters during the training stage. Left: \citet{houlsby2019parameter} add their Adapters after the feed-forward layers, and their Adapter consists of two linear layers and an active function. Middle: BitFit tunes all biases from the original transformer layers. Right: AdapterBias, consisting of a linear layer ($L_\alpha$) and a vector (\textit{v}), is added after the second feed-forward layer only in each transformer layer.}
      \label{fig:arch}
\end{figure*}

\section{Method}
%Traditional adapter approaches \citep{houlsby2019parameter} requires two feed-forward layers and an activation function. However, we think that it hasn't reached the simplest state. The concept of an adapter is to add a task-specific bias vector to the embedding of all layers. 

In this section, we present AdapterBias, an efficient way to adapt large-scale PLMs. 
In order to better adapt to different downstream tasks, the adapter module should be token-specific. AdapterBias produces a suitable weight for the bias based on the input token.

 %Take SST-2, a sentiment analysis dataset, for example. Focusing on adjective words, which contains more sentiment information, can help the pre-trained model to adapt to the task better. Therefore, adding a larger bias on the token of adjective word is more reasonable in this task. 

\textbf{Problem Formulation}
We consider the general problem of fine-tuning PLMs, where the training data $D={(x_i,y_i)}_{n=1}^N$ is given. %We assume we are also given a pre-trained language model $ f_{\theta}(.)$ parameterized by $\theta$ that computes the output for input $x_i$. 
Assume that given a PLM with parameters $\theta$ and AdapterBias with parameters $\theta'$. During the training stage, we freeze $\theta$ and tune $\theta'$ only. 

\subsection{AdapterBias}
%To adapt the PLM to the downstream tasks, our main concept is to add a token-specific embedding shift to the output of the pre-trained transformer layer through our AdapterBias.  hi

%To learn the embedding shift in a more parameter-efficient way, instead of using two linear layer to transform the output token \citep{houlsby2019parameter}   
The architecture of AdapterBias is shown in the right part of Figure \ref{fig:arch}. AdapterBias consists of two modules: a vector (\textit{v}) and a linear layer (\textit{$L_{\alpha}$}).
\textit{v} is a task-specific shift added to the output of each transformer layer. 
The tokens which are more related to the task should be assigned larger representation shifts than other tokens. The linear layer (\textit{$L_{\alpha}$}) produces a token-dependent weight vector $\alpha = \left[\alpha_1, \alpha_2 \dots \alpha_m\right]^T$, where $\alpha_i$ is the weight of the $i^{th}$ token's representation shift. 
By applying the token-specific weight to the task-specific representation shift (\textit{v}), AdapterBias can focus on the tokens that are more important to the task and is able to adapt to different downstream tasks efficiently.

%Our AdapterBias consists of only a vector (\textit{v}) and a linear layer (\textit{$L_{\alpha}$}).
%The vector is designed as input-independent which represents the characteristic of each transformer layer. Noted that $v \in \mathbb{R}^{e}$, and $e$ is the size of an embedding. Our AdapterBias have a linear layer (\textit{$L_{\alpha}$}) aiming to learn a single value, $\alpha$, which acts as the weight of the bias for different input tokens. For example, if we have a sentence with m tokens, the corresponding dimension of $\alpha$ will be $m \times 1$.

%In order to make the vector better adapt to the downstream task, our AdapterBias have a linear layer (\textit{$L_{\alpha}$}) aiming to learn weights of the bias for different tokens, so as to make it token specific. Noted that the output of \textit{$L_{\alpha}$} called $\alpha$, where $\alpha \in \mathbb{R}^{l}$ and \textit{l} represents the length of sentence.

We define the output of AdapterBias as the bias (\textit{B}), which is the outer product of \textit{v} and the learned weights vector $\alpha$. When the dimension of the token's representation is $r$ with $m$ input tokens, the function can be defined as follows: 
%The output of AdapterBias called the bias (\textit{B}), which is the outer product of \textit{v} and $\alpha$. The function can be defined as follows:
\begin{equation}
\normalsize B=   v \otimes\alpha^T =
%\begin{pmatrix}
%\alpha_{1}v_1 & \dots & %\alpha_{m}v_1\\
% \vdots & \ddots & \vdots\\
%\alpha_{1}v_e & \dots & \alpha_{m}v_e
%\end{pmatrix}_{e \times m}
\begin{pmatrix}
\alpha_{1}v & \alpha_{2}v &\dots &\alpha_{m}v
\end{pmatrix}
\label{eq:adpbias}
\end{equation}
where $v \in \mathbb{R}^{r}$, $\alpha \in \mathbb{R}^{m}$, and $B \in \mathbb{R}^{r \times m}$.

%As shown in Figure \ref{fig:arch}, $B$ is added to the output of the last feed-forward layer and the residual network. 
%The parameters from the second normalization layer in each transformer are also tuned during adapter-tuning.

\begin{figure}
    \centering
      \includegraphics[width=0.38\textwidth]{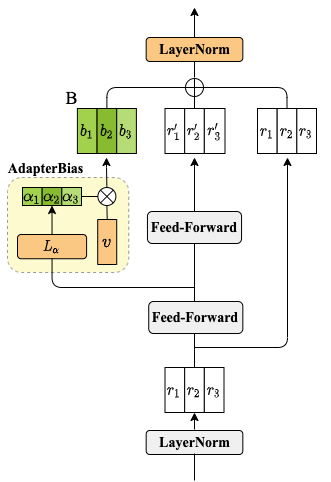}
      %\caption{model architecture of vector linear}
      \caption{The detailed architecture of how AdapterBias produces the bias (\textit{B}) and how \textit{B} is added to the output of transformer layers.}
      \label{fig:onlyour}
\end{figure}

To further elaborate on the details of AdapterBias, we give an example of how AdapterBias produces \textit{B} and how \textit{B} adds to the transformer layer. In Figure~\ref{fig:onlyour}, we assume that there are three representation outputs ($r_1, r_2, r_3$) after the first layer normalization. The dimension of $r_1$, $r_2$ and $r_3$ is the dimension of the 2\textsuperscript{nd} feedforward layer, while the input dimension of the linear layer ($L_\alpha$) is the output dimension of the first feed-forward layer with the token representation ($r_1, r_2, r_3$) as its inputs.
%The dimension of $r_1$, $r_2$ and $r_3$ is 768. Note that the dimension of the vector (\textit{v}) in AdapterBias is also 768 and the dimension of the linear layer ($L_\alpha$) is 3072, which is the output of the first feed-forward layer with inputting token representation ($r_1, r_2, r_3$).
The linear layer ($L_\alpha$) produces $\alpha$, where $\alpha \in \mathbb{R}^{3}$. The blocks in different colors represent the difference of the weights ($\alpha_1,\alpha_2,\alpha_3$). Take BERT-base for example, after performing outer product with the weights vector $\alpha$ and the vector (\textit{v}), the dimension of \textit{B} becomes $768 \times 3$. For example, $b_1$, the first column of \textit{B}, is the shift for the first token representation.

%Thus, we can define our trainable parameter $\theta'$, correspond to the orange components in right part of Figure \ref{fig:arch},
%\begin{equation}
%\theta' = \theta'_v + \theta'_{L_{\alpha}},
%\label{eq:2}
%\end{equation}
%where $\theta_{{LN}_2}, \theta^{AD}_v, \theta^{AD}_{L_{\alpha}}$ represent the parameters of the normalization layer in each transformer layer, the vector and linear layer in AdapterBias, respectively. Noted that $\theta_{{LN}_2}$ are the original PLM's parameters.

%By tuning AdapterBias, a token-specific embedding shift is added to each pre-trained transformer layer, which enables the pre-trained BERT to adapt to downstream tasks with extremely low trainable parameters.  

%\subsection{More parameter-efficient in AdapterBias}
\subsection{Further improvement on parameter-efficiency of AdapterBias}

In this section, we experiment on two different methods to make AdapterBias more parameter efficient. One is partial weight-sharing of AdapterBias among transformer layers, another is enforcing the weights of the linear layer ($L_\alpha$) to be sparse by utilizing $L_0$-norm penalty.

\subsubsection{Cross-layer parameters sharing in AdapterBias}

Redundancies have been observed in the information captured by adapters, with adapters in lower layers being less important  \cite{houlsby2019parameter}. In addition, sharing parameters of the Adapter across layers leads to a comparatively small drop in performance in some tasks.
In light of the above information, we further reduce the number of parameters required for each task by partially sharing the weights of the adapters across all transformer layers. The experimental results are discussed at Section~\ref{sec:exp_share}.

\subsubsection{$L_0$ regularization in AdapterBias}

Sparsity has been utilized in various parameter-efficient methods. For applications in NLP tasks, Diff-pruning~\citep{guo2020parameter} learns a sparse vector added to the whole PLM with $L_0$-norm penalty. Inspired by their work, we further apply $L_0$-norm regularization to $L_\alpha$ in the AdapterBias module, aiming to encourage the sparsity of $L_\alpha$. We choose to drop $L_\alpha$ because it contributes most of the parameters in AdapterBias. Encouraging its sparsity can further increase the parameter efficiency. 
Note that we specifically apply $L_0$ regularization in Section~\ref{sec:exp_L0}.

In AdapterBias, we add $L_0$-norm penalty to the linear layer ($L_\alpha$). 
The optimization problem can be expressed as,
\begin{equation}
\begin{aligned}
\normalsize
\mathop{min}\limits_{\theta'} L(D;\theta,\theta') + \lambda \| \theta'_{L_{\alpha}} \|_0,
\label{eq:6}
\end{aligned}
\end{equation}
where $L(D;\cdot)$ represents the original loss with training data $D$. $\lambda$ is the hyperparameter for $L_0$-norm penalty. Note that $\theta'$ represents trainable parameters and $\theta'_{L_\alpha}$ represents the parameters of $L_\alpha$ in AdapterBias. 
Following the work of Diff-pruning, we utilize a relaxed mask vector \cite{louizos2017learning} with a stretched Hard-Concrete distribution \cite{jang2016categorical, maddison2016concrete} to encourage $L_0$ sparsity. 

\begin{table*}[h]
\resizebox{\textwidth}{15.5mm}{
\begin{tabular}{c|l|c|cccccccccc}
\hline
\multicolumn{2}{c|}{\textbf{Method}}      & \textbf{Params} & \textbf{CoLA} & \textbf{SST-2} & \textbf{MRPC} & \textbf{QNLI} & \textbf{RTE} & \textbf{STS-B} & \textbf{MNLI-m} & \textbf{MNLI-mm} & \textbf{QQP} & \textbf{Avg} \\ \hline
\multicolumn{2}{c|}{BERT{$\rm _{LARGE}$}}             & 340M                & 60.5         & 94.9           & 89.3          & 92.7          & 70.1         & 87.6          & 86.7            & 85.9             & 72.1         & 82.2         \\
\multicolumn{2}{c|}{Adapters~\citep{houlsby2019parameter}}              & 7.14M               & 56.9          & 94.2           & 89.6          & 91.4          & 68.8         & 87.3           & 85.3            & 84.6             & 71.8         & 81.1         \\
% \multicolumn{2}{c|}{\citet{pfeiffer2020adapterfusion}}                  & 68M                 & 59.3          & 94.7           & 87.6          & 91.5          & 71.5         & 86.5           & 85.2            & 84.3             & 71.4         & 81.3         \\
\multicolumn{2}{c|}{Diff-Pruning~\citep{guo2020parameter}}                  & 1.7M                & 61.1          & 94.1           & 89.7          & 93.3          & 70.6         & 86.0           & 86.4            & 86.0             & 71.1         & 82.0         \\
\multicolumn{2}{c|}{BitFit~\citep{ben2021bitfit}}                  & 0.27M               & 59.7          & 94.1           & 88.9          & 92.0          & 72.0         & 85.5           & 84.5            & 84.8             & 70.5         & 81.3         \\
\multicolumn{2}{c|}{LoRA~\citep{hu2021lora}}                  & 0.39M               & 60.6          & 94.0           & 87.9          & 92.2          & 70.3         & 85.6           & 84.2            & 84.0             & 70.0         & 81.0         \\\hline\hline
\multicolumn{2}{c|}{\textbf{AdapterBias}} &  0.17M               & 60.0          & 94.4           & 88.2          & 91.2          & 70.5         & 87.5           & 84.3            & 83.9            & 70.5         & 81.2         \\\hline

%\multicolumn{2}{c|}{\textbf{AdapterBias++}} &  \textbf{0.042}\%               & 58.7          & 94.1           & 87.3          & 90.7          & 68.9         & 86.8           & 83.4            & 82.8            & 70.3         & 80.3         \\ \hline

\end{tabular}}
\caption{Performance of all methods on the GLUE testing sets scored by the GLUE evaluation server. For each method, we report the new adding parameters per task. For QQP, we report the F1 score. For STS-B \citep{cer2017semeval}, we report Spearman correlation coefficients. For CoLA \citep{warstadt2019neural}, we report Matthews correlation. For all other tasks, we report accuracy. Bold fonts indicate the least trainable parameter per task. The first row (BERT$\rm_{LARGE}$) represents fine-tuning the whole BERT-large model without adding new parameters. The results of baselines including \citep{houlsby2019parameter,guo2020parameter,ben2021bitfit} are their reported performance and \citet{pfeiffer2020adapterfusion, hu2021lora} performance is reproduced on our setting. Due to instability during training, we restart experiments with 3 random seeds and report the best. 
}
\label{tb:glue}
\end{table*}
\section{Experiments}
In this section, we evaluate the effectiveness of our proposed adapter module in NLP training tasks, and provide the analysis of what AdapterBias has learned in different tasks.
%thus requires extensive experiments.

%layernorm是有train的

\subsection{Experimental settings}
We base our experiments on HuggingFace PyTorch implementation \citep{wolf2019huggingface} of BERT \citep{devlin2018bert} and RoBERTa \citep{liu2019roberta} models. The learning rate is set in the range [$10^{-4}$, $10^{-3}$], with AdamW \citep{loshchilov2017decoupled} as the optimizer. GLUE benchmark \citep{wang2018glue} and SQuAD v1.0 \cite{rajpurkar2016squad} are the training data in our settings. 
%We train most tasks with a minibatch size of 32 and a maximum sequence length of 128 tokens. For STS-B \citep{cer2017semeval} and QNLI \citep{rajpurkar2016squad}, we set the maximum sequence length to 512 tokens with a minibatch size of 16.

The training details are shown in Appendix \ref{sec:appendix}. Note that the second layer normalization in each transformer layer is also tuned during the training stage, corresponding to the orange component in the right part of Figure \ref{fig:arch}. We experiment with 3 random seeds and choose the seed with the best performance on the validation set to evaluate on the GLUE server. We report the test metrics provided on the submission website\footnote{https://gluebenchmark.com/}.

%We choose the best model from within that training time based on the score of development set. No warmup step method is used in our training. For RTE \citep{bentivogli2009fifth}, we specify the token id due to the sensitivity of the dataset to token id.

\begin{table*}[]
\resizebox{\textwidth}{27mm}{
\begin{tabular}{clclccccccccccc}
\hline
\multicolumn{2}{l}{}    & \multicolumn{2}{c}{\textbf{Method}} & \textbf{Params} & \textbf{CoLA} & \textbf{SST-2} & \textbf{MRPC} & \textbf{QNLI} & \textbf{RTE} & \textbf{STS-B} & \textbf{MNLI-m} & \textbf{MNLI-mm} & \textbf{QQP} & \textbf{Avg} \\ \hline
\multicolumn{2}{c}{BB}  & \multicolumn{2}{c}{Full-FT}         & 110M             & 52.1          & 93.5           & 88.9          & 90.5          & 66.4         & 85.8           & 84.6            & 83.4             & 71.2         & 79.6         \\

\multicolumn{2}{c}{BB}  & \multicolumn{2}{c}{BitFit}     & 0.10M          & 47.2         & 92.4           & 87.4          & 89.7          & 65.5         & 87.6           & 80.8            & 80.9            & 67.8         &77.7\\
\multicolumn{2}{c}{BB}  & \multicolumn{2}{c}{AdapterBias}     & 0.06M          & 51.6          & 93.1           & 87.5          & 89.4          & 66.1         & 84.6           & 80.9            & 80.5             & 67.9         & 78.0

\\ \hline
\multicolumn{2}{c}{BL}  & \multicolumn{2}{c}{Full-FT}         & 340M             & 60.5          & 94.9           & 89.3          & 92.7          & 70.1         & 87.6           & 86.7            & 85.9             & 72.1         & 82.2         \\

\multicolumn{2}{c}{BL}  & \multicolumn{2}{c}{BitFit}     & 0.27M           & 62.0         & 93.1          & 86.8         & 89.8          & 66.6         & 87.2           & 84.1           & 	84.3             & 67.2         & 80.1        \\

\multicolumn{2}{c}{BL}  & \multicolumn{2}{c}{AdapterBias}     & 0.17M          & 60.0          & 94.4           & 88.2          & 91.2          & 70.5         & 87.5           & 84.3            & 83.9             & 70.5         & 81.2         \\ 

\hline
\multicolumn{2}{c}{RoB} & \multicolumn{2}{c}{Full-FT}         & 125M             & 61.3          & 94.7           & 90.4          & 92.0          & 74.4         & 87.5           & 87.4            & 86.8             & 71.9             & 82.9             \\

\multicolumn{2}{c}{RoB}  & \multicolumn{2}{c}{BitFit}     & 0.10M           & 62.7          & 94.8           & 89.7          & 91.3          & 73.6         & 88.5          & 85.3            & 84.9             & 68.1         & 82.1        \\

\multicolumn{2}{c}{RoB} & \multicolumn{2}{c}{AdapterBias}     & 0.06M           & 61.9          & 94.5           & 90.2          & 91.1          & 74.1         & 88.7           & 85.3            & 85.1             & 70.5         & 82.4

\\

\hline
\multicolumn{2}{c}{RoL} & \multicolumn{2}{c}{Full-FT}         & 355M             & 63.3              & 96.7               & 92.3              & 95.4              & 84.5             & 92.2               & 90.8                & 90.2                 & 74.3             & 86.6             \\

\multicolumn{2}{c}{RoL}  & \multicolumn{2}{c}{BitFit}     & 0.26M           & 64.7         & 95.8           & 91.5          & 94.2          & 80.9        & 90.6           & 89            & 88.9           & 72.0        & 85.3         \\

\multicolumn{2}{c}{RoL} & \multicolumn{2}{c}{AdapterBias}     & 0.17M           & 63.9          & 96.4           & 90.4          & 94.7          & 83.6         & 91.3           & 89.8            & 89.4             &  72.3            & 85.8             \\

\hline
\end{tabular}}
\caption{
Performance of AdapterBias adding in different PLMs. Here we experiment with four models : BERT-base (BB), BERT-large (BL), RoBERTa-base (RoB), and RoBERTa-large (RoL). The settings are the same as in Table \ref{tb:glue}. The Full-FT corresponds to fine-tuning the whole PLM without adding adapters.
}
\label{tb:roberta}
\end{table*}
\begin{table*}[h]
\resizebox{\textwidth}{8mm}{
\begin{tabular}{c|l|c|cccccccccc}
\hline
\multicolumn{2}{c|}{\textbf{Method}}      & \textbf{Params} & \textbf{CoLA} & \textbf{SST-2} & \textbf{MRPC} & \textbf{QNLI} & \textbf{RTE} & \textbf{STS-B} & \textbf{MNLI-m} & \textbf{MNLI-mm} & \textbf{QQP} & \textbf{Avg} \\ \hline
\multicolumn{2}{c|}{w/o \textit{$L_\alpha$}}     & 27.6K               & 45.6          & 91.5           & 87.4          & 88.3          & 65.6         & 81.0           & 77.9            & 78.4             & 65.7         & 75.7         \\
 \hline\hline
\multicolumn{2}{c|}{AdapterBias} & 64.5K              & \textbf{51.6}          & \textbf{93.1}           & \textbf{87.5}          & \textbf{89.4}          & \textbf{66.1}         & \textbf{84.6}           & \textbf{80.9}            & \textbf{80.5}             & \textbf{67.9}         & \textbf{78.0}         \\ \hline
\end{tabular}}
\caption{Evaluating the importance of the linear layer ($L_\alpha$) in AdapterBias. The settings are the same as in Table \ref{tb:glue}. The backbone model is BERT-base. w/o $L_\alpha$ means that there is only a vector (\textit{v}) in AdapterBias.}
\label{tb:ablation}
\end{table*}
\subsection{Results on GLUE}
%%%%%%%%%%%%%%%%%%%TODO%%%%%%%%%%%%%%%%%%%

%(the number of parameters in the BERT-base model is considered as $1 \times$)
%Although this approach achieves better performance compared with all the adapter-based models, it fine-tunes all the parameters in the BERT model in each task, and thus it requires 9x of the parameters than BERT-base in nine tasks of GLUE benchmark
In this section, we compare AdapterBias to other parameter-efficient methods, including Adapters \citep{houlsby2019parameter}, Diff-pruning \citep{guo2020parameter}, BitFit \citep{ben2021bitfit}, and LoRA \citep{hu2021lora}. 
In Table \ref{tb:glue}, we report the test scores on the GLUE benchmark and the required new parameters per task. Here we use BERT-large as the PLM. AdapterBias reaches 81.2 average score in GLUE benchmark, with the smallest amount of parameters (0.17M) added per task. AdapterBias shows competitive performance as its parameters are 40$\times$ less than the works of \citet{houlsby2019parameter}.
Although Diff-pruning \citep{guo2020parameter} achieves the best average score among all parameter-efficient methods, their work trains an additional vector whose parameter count is equivalent to the parameters of the whole PLM. Thus, Diff-pruning requires 340M trainable parameters of BERT-large during the training stage, while AdapterBias only trains 0.17M parameters.
Furthermore, AdapterBias achieves comparable performance with BitFit and LoRA with fewer parameters needed per task. This shows that AdapterBias is a worthwhile targeted fine-tuning method.

%; AdapterBias$_{learn}$ can reach 80.19 of average score in GLUE server, using with $1.004\times$ of the whole BERT-base parameters. Overall, our proposed AdapterBias requires $0.02\times$ of trainable parameters than that of \citet{houlsby2019parameter} while achieving 5\% higher accuracy than their work.

\subsection{Different base models}
To analyze the generalization ability of this approach to different PLMs on different models of AdapterBias, as shown in Table \ref{tb:roberta}, we apply AdapterBias in different transformer-based PLMs, including BERT-base (BB), BERT-large (BL), RoBERTa-base (RoB), and RoBERTa-large (RoL), on the GLUE benchmark. All results are scored by the GLUE evaluation server. Compared with BitFit,
In Table \ref{tb:roberta}, not only can AdapterBias perform well on BERT but also achieve competitive performance on larger PLMs such as RoBERTa.

\subsection{Size of training data}
\begin{figure}
    \centering
      \includegraphics[width=0.5\textwidth]{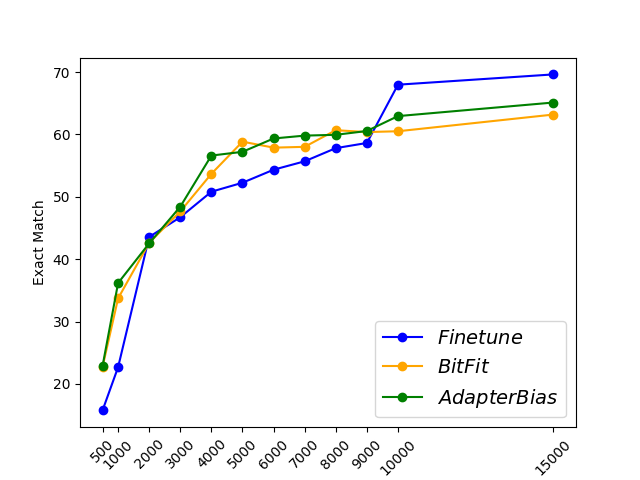}
      %\caption{model architecture of vector linear}
      \caption{Comparison of Finetune, BitFit \cite{ben2021bitfit}, and AdapterBias with BERT-base on SQuAD validation set. The x-axis represents the total number of training examples while the y-axis represents the exact match score.}
      \label{fig:fewshot}
\end{figure}

%%%%%%%%%%%%%%%%%%%TODO%%%%%%%%%%%%%%%%%%%%%%
In the previous experimental results, we observe that AdapterBias tends to have higher performance on tasks with a smaller amount of data (i.e. CoLA, SST-2, and RTE). To further validate this observation, we follow the work of BitFit \citep{ben2021bitfit} by training AdapterBias on subsets of SQuAD v1.0 \citep{rajpurkar2016squad} of increasing size. The experiments are conducted with BERT-base. The results on the validation set of the SQuAD dataset are listed in Figure \ref{fig:fewshot}, which shows the tendency of AdapterBias outperforming full fine-tuning when the size of the training dataset is smaller. However, with more training data available, the trend is reversed. The results show that AdapterBias has the ability to outperform fine-tuning the whole PLM with small-to-medium data size, similarly to BitFit. 
%not only does AdapterBias have comparable performance with BitFit with less trainable parameters required, it can also outperform fine-tuning the whole PLM when small-to-medium data size.

% We compare $Adapter_\alpha$ of different bottlenecks (blue) with $Adapter_\xi$ of different size of bottlenecks (orange). And we use fine-tune average score as standard line.

%\input{Figures/parameter_trade_off}

%To compare our experimental results to other adapter approaches, we reproduce \citet{houlsby2019parameter} on BERT-base with bottleneck set to 64. All the bottlenecks of our adapters ($Adapter_{\alpha}$ and $Adapter_{\xi}$) are fixed to 8

%on of BitFit and Full-FT with BERTBASE exact match score on SQuAD validation set.

\subsection{Investigation on the effectiveness of token dependent representation shift} %vector中的參數能不能drop掉
Different from BitFit~\citep{ben2021bitfit}, where the bias terms in all transformer layers are tuned, we claim that the bias added to the representation should be token-dependent, and proposed AdapterBias based on this concept.  We conduct ablation studies to verify this claim. In this experiment, the linear layer ($L_\alpha$) in AdapterBias that produces the token-dependent weights vector ($\alpha$) is removed; that is, only the \textit{v} is trained. All shifts added to the representation outputs are identical within the same transformer layer. The experiments are conducted with BERT-base model. We report the test scores on the GLUE benchmark in Table~\ref{tb:ablation}. The performance of AdapterBias without the linear layer ($L_\alpha$) dramatically decreases. Without $L_{\alpha}$, it is hard for the vector (\textit{v}) to adapt to different downstream tasks. This result demonstrates the importance of $L_{\alpha}$. In other words, assigning different shifts to different token representations improves the performance of the method.

%In this section, experiments are conducted to demonstrate the importance of $\alpha$, which is the weight of the bias. We only train the vector component by removing the linear layer that produces $\alpha$, and report the validation score in Glue benchmarks.

%\subsubsection{AdapterBias without $L_\alpha$} % vector only
%Different from BitFit where the bias term in every transformer layer is tuned, AdapterBias adds a token dependent bias shift to the token by adding a linear layer ($L_\alpha$) to produce the weight to the vector. To demonstrate the importance of $\alpha$, we remove $L_\alpha$, so that all the shifts added to the token are the same in a task. Table \ref{tb:ablation} shows the experimental result on Glue benchmark conducted with BERT-base model. As shown in Table \ref{tb:ablation}, the result indicates that with the linear layer ($L_\alpha$), AdapterBias outperforms training the vector only since the same bias would be added to each token in only vector method. On the other hand, with $L_\alpha$, bias for each token depends on $\alpha$, the output of $L_\alpha$. This verifies the importance of adding different bias shift based on the token.
%Therefore, adding different bias to different tokens is essential.     hello
\begin{table*}[h]
\resizebox{\textwidth}{12mm}{
\begin{tabular}{c|l|c|cccccccccc}
\hline
\multicolumn{2}{c|}{\textbf{Method}}      & \textbf{Params} & \textbf{CoLA} & \textbf{SST-2} & \textbf{MRPC} & \textbf{QNLI} & \textbf{RTE} & \textbf{STS-B} & \textbf{MNLI-m} & \textbf{MNLI-mm} & \textbf{QQP} & \textbf{Avg} \\ \hline
\multicolumn{2}{c|}{Share \textit{v}}          & 56.1K              & 50.1          & 90.8           & 87.1          & 87.6          & 65.0         & 84.9           & 77.5            & 77.9             & 65.1         & 76.2         \\
\multicolumn{2}{c|}{Share $L_\alpha$}            & 30.7K              & 50.4          & 91.9           & \textbf{88.1}          & 89.1          & 65.4         & 85.2           & 79.8            & 79.9             & 66.6         & 77.4         \\
\multicolumn{2}{c|}{Share \textit{v+}$L_\alpha$}    & 22.3K   & 46.8          & 90.9           & 87.3          & 87.8          & 64.8         & \textbf{85.7}          & 77.7            & 78.0             & 64.9         & 76.0        \\ \hline\hline

%\multicolumn{2}{c|}{ AdapterBias (L0)}     & 0.062\%               & \textbf{53.7}          & 92.5           & 87.5          & \textbf{90.3}          & \textbf{68.3}         & 85.7           & \textbf{81.7}            & \textbf{81.5}             & \textbf{69.8}         & \textbf{79.0}         \\ \hline\hline
\multicolumn{2}{c|}{\textbf{AdapterBias}} & 64.5K              & 51.6         & \textbf{93.1}           & 87.5          & 89.4          & 66.1         & 84.6           & 80.9            & 80.5             & 67.9         & 78.0         \\ \hline
\end{tabular}}
\caption{ Analysis of more parameter-efficiency methods in AdapterBias. The settings are the same as in Table \ref{tb:glue}. The backbone model is BERT-base. Share \textit{v}, Share $L_\alpha$, and Share \textit{v+}$L_\alpha$ means that we share vector, linear layer, and both of them, respectively. }
\label{tb:efficient}
\end{table*}
\subsection{Improving the parameter efficiency of AdapterBias}
%Although AdapterBias already uses fewer new parameters, 

We further apply two additional methods to AdapterBias to enhance its parameter efficiency. Experiments are conducted to exami whether AdapterBias can be more parameter-efficient by sharing its components across all layers. Moreover, we experiment on adding $L_0$-norm regularization during the training stage to encourage the sparsity of AdapterBias.

%t on whether AdapterBias can be more parameter efficient by sharing different components of AdapterBias module across all layers, that is, the linear layer that produces $\alpha$ and vector.
%Furthermore, we experiment on adding $L_0$-norm regularization to learn sparse vectors.
\begin{table*}[h]
\resizebox{\textwidth}{15mm}{
\begin{tabular}{clclcccccccccc}
\hline
\multicolumn{2}{c}{}   & \multicolumn{2}{c}{\textbf{Method}}  & \textbf{CoLA} & \textbf{SST-2} & \textbf{MRPC} & \textbf{QNLI} & \textbf{RTE} & \textbf{STS-B} & \textbf{MNLI-m} & \textbf{MNLI-mm} & \textbf{QQP} & \textbf{Avg} \\ \hline
\multicolumn{2}{c}{BB} & \multicolumn{2}{c}{Full-FT}          & 52.1          & 93.5           & 88.9          & 90.5          & 66.4         & 85.8           & 84.6            & 83.4             & 71.2         & 79.6         \\
\multicolumn{2}{c}{BB} & \multicolumn{2}{c}{AdapterBias}      & 51.6          & 93.1           & 87.5          & 89.4          & 66.1         & 84.6           & 80.9            & 80.5             & 67.9         & 78.0         \\
\multicolumn{2}{c}{BB} & \multicolumn{2}{c}{AdapterBias ($L_0$)} & 53.7          & 92.5           & 87.5          & 90.3          & 68.3         & 85.7           & 81.7            & 81.5             & 69.8         & 79.0         \\ \hline\hline
\multicolumn{2}{c}{BL} & \multicolumn{2}{c}{Full-FT}          & 60.5          & 94.9           & 89.3          & 92.7          & 70.1         & 87.6           & 86.7            & 85.9             & 72.1         & 82.2         \\
\multicolumn{2}{c}{BL} & \multicolumn{2}{c}{AdapterBias}      & 60.0          & 94.4           & 88.2          & 91.2          & 70.5         & 87.5           & 84.3            & 83.9             & 70.5         & 81.2         \\
\multicolumn{2}{c}{BL} & \multicolumn{2}{c}{AdapterBias ($L_0$)} & 58.0          & 93.7           & 88.2          & 91.5          & 69.2         & 87.2           & 84.2            & 84.1             & 71.2         & 80.8         \\ \hline
\end{tabular}}
\caption{
Performance of our AdapterBias with $L_0$-norm regularization. Here we experiment with two models: BERT-base (BB), and BERT-large (BL). The settings are the same as in Table \ref{tb:glue}. The Full-FT represents fine-tuning the whole PLM without adding adapters.
}
\label{tb:l0}
\end{table*}
\subsubsection{Sharing components in AdapterBias}
\label{sec:exp_share}

%Redundancies have been observed in the information captured by adapters, with adapters in lower layers being less important. In particular, sharing parameters of the adapter across layers leads to a comparatively small drop in performance for some tasks \citep{mahabadi2021compacter}. 
In this experiment, we conduct an ablation study of partial weight-sharing in the AdapterBias module. In Table \ref{tb:efficient}, we share components of AdapterBias among different transformer layers. \textit{Share} \textit{v} represents sharing \textit{v} across all transformer layers, while \textit{Share} $L_\alpha$ means sharing the linear layer ($L_\alpha$). \textit{Share} \textit{v+}$L_\alpha$ denotes sharing one AdapterBias across all transformer layers. As can be seen in Table \ref{tb:efficient}, the performance of \textit{Share} $L_\alpha$ stands out among other partial weight-sharing methods, while \textit{Share} \textit{v} leads to a poor performance.

From the experiments above, we conclude that the linear layer ($L_\alpha$) captures general task information by learning the weights of the bias for different tokens. Thus, sharing $L_\alpha$ across all layers results in better performance compared to other components. The vector module (\textit{v}) in AdapterBias aims to learn local information in each transformer layer. If \textit{v} among different transformer layers are shared, the performance drops dramatically. This might be due to a failure of \textit{v} to learn general information which can be adapted to each individual transformer layer.

%By inputting the sentences of same task, $L_\alpha$ is learned to be a task-related module so it can be shared among layers.
%On the other hand, \textit{Share} $v$ performs worst and this is because the vector ($v$) depends on transformer layer, and thus it cannot be shared among layers.
%If each layer is shifted by same bias, the model will not identify which layer is more important. 

%\subsubsection{Sparse adapter}

%If we can regularize our vector to be sparse such that kδτ k0  kθk0, then this approach can become more parameter-efficient as

% instead of

\subsubsection{$L_0$-norm regularization in AdapterBias}
\label{sec:exp_L0}

We observed that many of the trained parameters in $L_\alpha$ have values that are extremely close to zero after tuning on downstream tasks, which might cause redundancy of the parameters. To further encourage the sparsity of AdapterBias, we add $L_0$-norm regularization to $L_\alpha$ during the training stage.

In Table \ref{tb:l0}, we use BERT-base (BB) and BERT-large (BL) as the PLMs. We compare the performance of fine-tuning, the original AdapterBias, and the one trained with $L_0$-norm regularization. The experiment shows that adding $L_0$-norm regularization during the training step improves the performance on 7 out of 9 tasks in BERT-base models. However, the performance did not improve when applied to BERT-large models.
%and the method, $L_0$-norm performs better than the original AdapterBias.
As for the parameter efficiency of applying $L_0$-norm penalty, the linear layer ($L_\alpha$) with $L_0$-norm penalty saves about 17\% parameter on average compared to the original AdapterBias. The details of the reduced parameters of each task are shown in Appendix \ref{sec:appendix}.
%The result shows that our AdapterBias can further drop almost 17\% parameters with containing comparable performances in Glue benchmark. The detail saving parameter for each task is shown in Appendix.

%By this study, it is surprising that we can diminish our adapter parameters and even have higher performance on about 20\% dropping ratio. We further conclude that it is because some parameters are relatively small and it became some noise to our adapter.
%\input{Figures/sparse}
%\input{Figures/sparse_4}
%\input{Figures/drop}

%Does alpha really learns the weight of the token embedding shift?
%Deeper look on what AdapterBias have learned

\subsection{What AdapterBias learns}
AdapterBias has good interpretability due to its simplicity. 
%One can see what AdapterBias focuses on when adapting downstream tasks from its token-dependent based embedding shift.
Compared to the similar work BitFit~\citep{ben2021bitfit}, where the shifts are identical for all tokens, AdapterBias adds token-dependent shifts to the output representation. By observing these token-dependent shifts, we analyze what AdapterBias learns when adapting to downstream tasks. 
%Firstly, to see the embedding shift of each token after the transform of AdapterBias, PCA \citep{jolliffe2002springer} is used for dimension reduction. Moreover, to take a deeper look on which kind of token that needs a larger embedding shift, we focus on $\alpha$, which is the output of $L_{\alpha}$, as it represents the weight of the bias for each token. 
%Observing the values of $\alpha$ gives us a picture of which layer of the transformer output needs a larger embedding shift and what tokens is more crucial to AdapterBias when adapting to downstream tasks.
%Therefore, we analyze $\alpha$ from two different aspects: which transformer layer has the largest $\alpha$ and what kind of token does $\alpha$ focus on.

%1.看alpha在哪幾層transformer最大 ＝＝>哪幾層transformer最需要移動
%2. 在同一個task裏面alpha focus在哪一格token上
%3. 在不同task上adapter focus在什麼token上

%Noted that $\alpha$, the output of linear layer, represents the weight of bias for each token and therefor we utilize the value of $\alpha$ to observe what tokens does AdapterBias focus on. For a token, we average all layers of $\alpha$. Then, we use the average weight in wordcloud. We experiment on two dataset : CoLA \citep{warstadt2019neural}, SST-2 \citep{socher2013recursive}. We use BERT-base pretrained in this section. The results show that our AdapterBias concentrates on tokens, which is important to tasks, by adding more large bias. 
\begin{figure}
    \centering
      \includegraphics[width=0.49\textwidth]{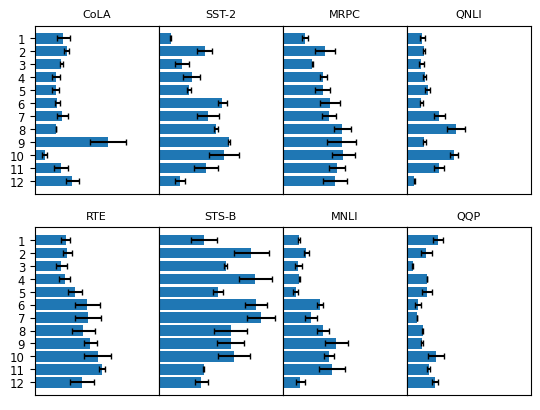}
      \caption{ 
      We analyze the average absolute value of weights vector $\alpha$, the output of the linear layer ($L_\alpha$), in each layer for different tasks.
      The y-axis represents the index of transformer layers, ordered from earlier to later (i.e. the embedding layer is shown at the top). The x-axis represents the average absolute value of $\alpha$.}
      \label{fig:layer_mean}
\end{figure}

%\subsubsection{The direction of embedding shifts in different tasks }
%Different from BitFit~\citep{ben2021bitfit}, where all the embedding shifts are identical within one task, AdapterBias produces different weights for the shift based on each token.
%In this section, we compare the transformed tokens in AdapterBias and BitFit. We utilize PCA~\citep{jolliffe2002springer} to reduce the dimension of the vectors. In Figure \ref{fig:PCA}, we input five sentences from the evaluation set of SST-2. We experiment on the last transformer layer since it has the most obvious shifts compared to the previous layers. '0' with lighter color indicates the embedding before shifting, which is the output of the first layer normalization. '1' with darker color is the shifted embedding, which is the output of the second layer normalization. The color red represents positive sentences, and blue are the negative ones.

%The result shows that BitFit shifts all tokens towards the same direction regardless of the ground-truth label. On the other hand, AdapterBias discerns the label of the sentences and thus shifts the tokens of different sentences toward different directions. 
%Furthermore, the special token, [CLS], has the largest change due to the fact that [CLS] is used to classify the sentence. Therefore, they need the largest shift to control the prediction.
\begin{figure}
     \centering
      \includegraphics[width=0.45\textwidth]{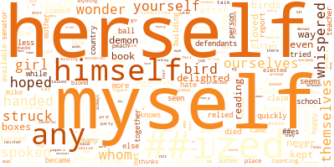}
      %\caption{model architecture of vector linear}
      \caption{Word cloud of CoLA, a corpus of linguistic acceptability. We utilize BERT-base model as the PLM and words come from validation data. The weights of the words are the summation of their weights produced by the linear layer ($L_\alpha$) in twelve transformer layers.}
      \label{wc:cola}
\end{figure}

%\subsubsection{Value of $\alpha$ in transformer layers}
\subsubsection{Average representation shifting in transformer layers}
\label{sec:alpha}
% compute的意義
%we want to know the average offset of the bias for tokens in each transformer layer. According to our designed, the linear layer ($L_\alpha$) produce the weights of embedding shift. Thus, we sum up all token's absolute weights of the shift in each transformer layer and then compute the average weight.

%Different layers of pretrained models have been
%argued to encode different informationn \citep{liu2019linguistic,tenney2019bert}. Given that each task is likely produce different kinds of language phenomena embedded in the hidden layers, we hypothesize that AdapterBias will modify different parts of the PLM through task-specific fine-tuning.

In light of the works of \citet{liu2019linguistic,tenney2019bert, kovaleva2019revealing}, which show that different information is being encoded by different transformer layers of PLMs. We assume that AdapterBias provides different representation shifts to the transformer layers through task-specific fine-tuning. 

In AdapterBias, the linear layer ($L_\alpha$) produces a weights vector $\alpha$ for representation shifts, therefore, the average absolute value of vector $\alpha$ can give us a look at the shifting amount in the transformer layers when adapting to downstream tasks. In Figure \ref{fig:layer_mean}, the layers are ordered from lower to upper. From the experimental result, we find that the weight in each layer is considerably different in different tasks in general. 

CoLA~\citep{warstadt2019neural} is a syntactic task that consists of English acceptability judgments in the GLUE benchmark. As shown in Figure~\ref{fig:layer_mean}, its average shift at the ninth layer is the highest among all layers, which is quite different from the others. We speculate that the ninth layer has the ability to extract the syntactic information, leading AdapterBias to add the largest shift in this layer. Our experiment has a similar observation with the work of \citet{jawahar2019does}. They observe on a syntactic task with BShift~\citep{conneau2018you} that the ninth layer of BERT embeds a rich hierarchy of syntactic information. \citep{jawahar2019does} 

Moreover, we observe similar distributions between specific tasks. For instance, RTE~\citep{giampiccolo2007third,bentivogli2009fifth} and MNLI~\citep{williams2017broad}, where both recognize textual entailment, have higher values in the upper layers than the lower ones. 

%MRPC and STS-B, which are both tasks detecting semantic similarity between two sentences, have relatively higher weights in all transformer layers compared to other tasks.

%In some tasks (i.e. SST-2, MRPC, and RTE), the value of $\alpha$ is similar in each layer. In STS-B, the value of $\alpha$ in the lower layer is higher than that in the upper layer. 
%In CoLA, the value of $\alpha$ is the highest at the ninth layer.
%In the rest of the tasks (i.e. QNLI, MNLI, and QQP), where the tasks encode questions, the value of $\alpha$ in the upper layer is higher than that in the lower layer. 

Based on these findings, we find that AdapterBias assigns suitable representation shifts in different tasks. For tasks with similar objectives, AdapterBias tends to add similar representation shifts.
 
\subsubsection{Which kind of word does $L_\alpha$ focus on}

%In our AdapterBias, the linear layer ($L_\alpha$) produce $\alpha$, which represent the weights of the embedding bias for each token. In light of this, we can utilize $\alpha$ to observe which token has a relative large shift. 

Since $\alpha_i$ represents the weight of the representation shift for $i^{th}$ token in a transformer layer, we can observe the significance of $i^{th}$ token from the summation of $\alpha_i$ in all the transformer layers. Special tokens, including [CLS], [SEP], and [PAD], are not included for analysis. We use the validation sets of CoLA and SST-2, and word cloud is used for visualizations.

In Figure \ref{wc:cola}, we visualize all words in the validation data of CoLA. The result shows that AdapterBias focuses more on reflexive pronouns, such as yourself, himself, and myself. This is because there are many incorrect sentences with misused reflexive pronouns, such as "He washed yourself." 

\begin{figure}
     \centering
      \includegraphics[width=0.45\textwidth]{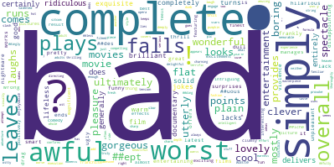}
      %\caption{model architecture of vector linear}
      \caption{Word cloud of SST-2, a corpus of movie reviews categorized in two sentimental classes (i.e. positive, negative). The visualization approach is the same as in Figure \ref{wc:cola}.}
      \label{wc:sst}
\end{figure}

In Figure \ref{wc:sst}, we visualize all words in the validation data of SST-2. The result shows that AdapterBias focuses more on adjectives, such as "bad", "awful", and "worst". SST-2 is a binary sentiment analysis dataset, which classifies movie reviews into positive and negative classes. AdapterBias learns that adjectives often constitute a crucial factor in sentiment analysis during tuning, and adds larger shifts to these adjective tokens. 

\section{Conclusion}
In this study, we present AdapterBias. By adding token-dependent representation shifts to the PLM, AdapterBias shows competitive results even though it uses far fewer parameters than the existing methods. Through extensive experiments, not only does AdapterBias reach competitive results on the GLUE benchmark, but also obtain good performance on small-to-medium datasets. In addition, we demonstrate the robustness of AdapterBias to different PLMs. 
Finally, we provide analysis on what AdapterBias learns by comparing $\alpha$, the weights of representation shift for different tokens, finding AdapterBias has the ability to identify task-specific information. 
Our study is different from the previous architectures of adapters by proposing a simple adapter that can produce suitable representation shifts for different tokens.

%overturn 感覺有點強XD

%yet powerful architecture for adapters, suitable for parameter-efficient transfer learning in NLP.

\section*{Acknowledgements}
We thank the National Center for High-performance Computing (NCHC) of National Applied Research
Laboratories (NARLabs) in Taiwan for providing computational and storage resources, and Chi-Liang Liu for giving us useful comments.
\bibliography{anthology,custom,ref}
\bibliographystyle{acl_natbib}

\appendix
\clearpage
\section{Appendix}

\setcounter{table}{0}
\setcounter{figure}{0}
\renewcommand{\thesection}{\Alph{section}}
\renewcommand{\thefigure}{\Alph{figure}}
\renewcommand{\thetable}{\Alph{table}}
\begin{table*}[h]
\resizebox{\textwidth}{11mm}{
\begin{tabular}{clccccccccc}
\hline
\multicolumn{2}{c}{}              & \textbf{CoLA} & \textbf{SST-2} & \textbf{MRPC} & \textbf{QNLI} & \textbf{RTE} & \textbf{STS-B} & \textbf{MNLI-m} & \textbf{MNLI-mm} & \textbf{QQP} \\ \hline
\multicolumn{2}{c}{Max\_len}      & 128           & 128            & 128           & 512           & 350          & 512            & 128             & 128              & 350          \\
\multicolumn{2}{c}{Batchsize}     & 32            & 32             & 32            & 16            & 32           & 16             & 32              & 32               & 32           \\ 
\multicolumn{2}{c}{Learning rate} & $10^{-3}$         & $10^{-3}$          & $10^{-3}$         & $10^{-4}$        & $4\times10^{-4}$       & $10^{-3}$          & $4\times10^{-4}$          & $4\times10^{-4}$           & $4\times10^{-4}$       \\
\multicolumn{2}{c}{Epoch}         & 20            & 10             & 10            & 10            & 20           & 20             & 10              & 10               & 10     \\\hline     
\end{tabular}}
\caption{
Our training details of GLUE benchmark\citep{wang2018glue}.
}
\label{app:1}
\end{table*}
\subsection{Training Details}
We train our model on Pytorch. The training details are shown in Table~\ref{app:1}. In addition, the bottleneck of Adapters~\citep{houlsby2019parameter} and 
\begin{table*}[]
\resizebox{\textwidth}{7mm}{
\begin{tabular}{clclccccccccc}
\hline
\multicolumn{2}{c}{}   & \multicolumn{2}{c}{\textbf{Method}}  & \textbf{CoLA} & \textbf{SST-2} & \textbf{MRPC} & \textbf{QNLI} & \textbf{RTE} & \textbf{STS-B} & \textbf{MNLI-m} & \textbf{MNLI-mm} & \textbf{QQP} \\ \hline
\multicolumn{2}{c}{BB} & \multicolumn{2}{c}{AdapterBias (L0)} & 26.2\%        & 82.0\%         & 83.1\%        & 82.3\%        & 81.0\%       & 83.0\%         & 83.2\%          & 83.3\%           & 83.4\%       \\
\multicolumn{2}{c}{BL} & \multicolumn{2}{c}{AdapterBias (L0)} & 83.2\%        & 83.0\%         & 83.3\%        & 83.7\%        & 83.2\%       & 83.2\%         & 83.4\%          & 83.7\%           & 83.6\%       \\ \hline
\end{tabular}}
\caption{
Percentage of remaining parameters compared with the original parameters of the linear layer ($L_\alpha$). Here we experiment with two models: BERT-base (BB) and BERT-large (BL). The setting follows by Table \ref{tb:glue}.
}
\label{app:2}
\end{table*} is 32.

\subsection{$L_0$-norm regularization in AdapterBias}
In Table~\ref{app:2}, we report the remaining parameters of utilizing $L_0$-norm regularization compared with the original AdapterBias. BERT-base (BB) and BERT-large (BL) are used as PLMs.
\subsection{The direction of representation shifts in different tasks}
\begin{figure}[t]
    \flushright
      \includegraphics[width=0.48\textwidth]{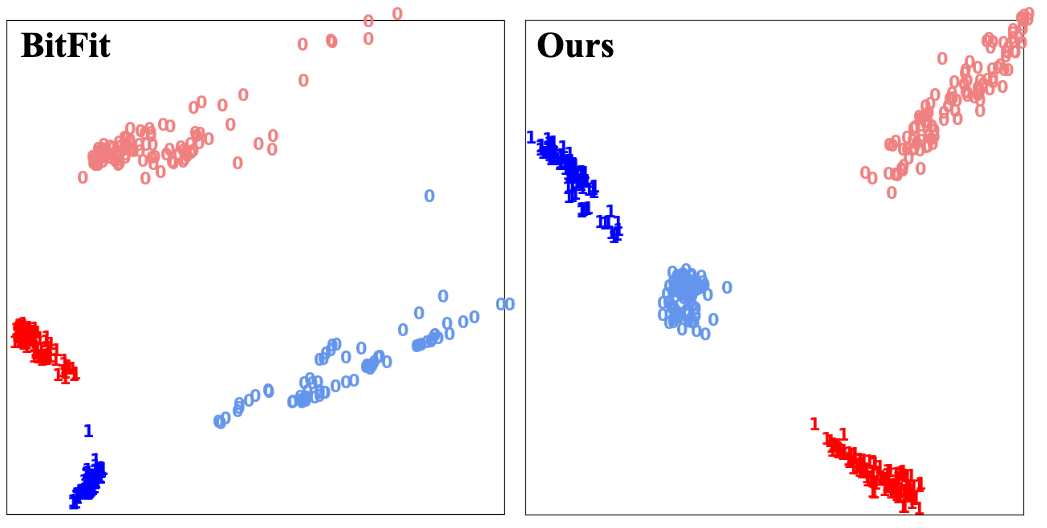}
      %\caption{model architecture of vector linear}
      \caption{We utilize PCA \citep{jolliffe2002springer} to visualize the shifting difference between Bitfit \cite{ben2021bitfit} and AdapterBias on SST-2 validation set. '0' with light color means the embedding before shifting. '1' with dark color means the embedding after shifting. The color red represents positive sentences, and blue represents negative sentences.}
      \label{fig:PCA}
\end{figure}

Different from BitFit~\citep{ben2021bitfit}, where all the representation shifts are identical within one task, AdapterBias produces different weights for the shift based on each token.
In this section, we compare the transformed tokens in AdapterBias and BitFit. We utilize PCA~\citep{jolliffe2002springer} to reduce the dimension of the vectors. In Figure \ref{fig:PCA}, we input five sentences from the evaluation set of SST-2. We experiment on the last transformer layer since it has the most obvious shifts compared to the previous layers. '0' with lighter color indicates the representation before shifting, which is the output of the first layer normalization. '1' with darker color is the shifted representation, which is the output of the second layer normalization. The color red represents positive sentences, and blue are the negative ones.

The result shows that BitFit shifts all tokens towards the same direction regardless of the ground-truth label. On the other hand, AdapterBias discerns the label of the sentences and thus shifts the tokens of different sentences toward different directions.

\begin{figure}
     \centering
      \includegraphics[width=0.4\textwidth]{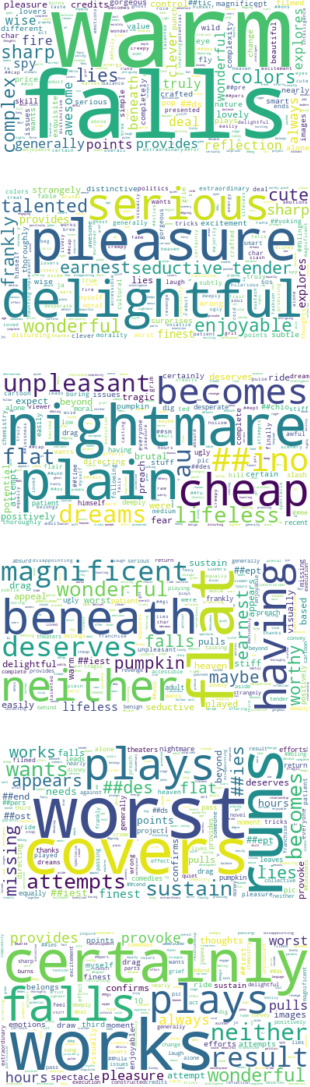}
      %\caption{model architecture of vector linear}
      \caption{Word cloud of SST-2 in layer 0 to layer 6.}
      \label{wc:sst_0-6}
\end{figure}

\begin{figure}
     \centering
      \includegraphics[width=0.4\textwidth]{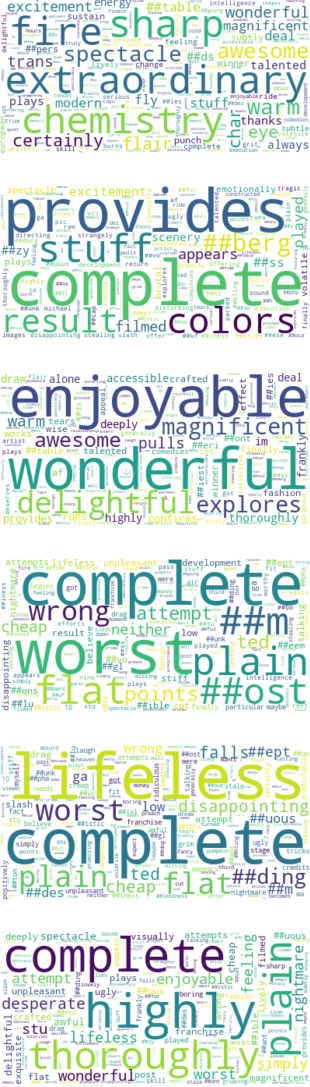}
      %\caption{model architecture of vector linear}
      \caption{Word cloud of SST-2 in layer 7 to layer 12.}
      \label{wc:sst_7-12}
\end{figure}

\begin{figure}
     \centering
      \includegraphics[width=0.4\textwidth]{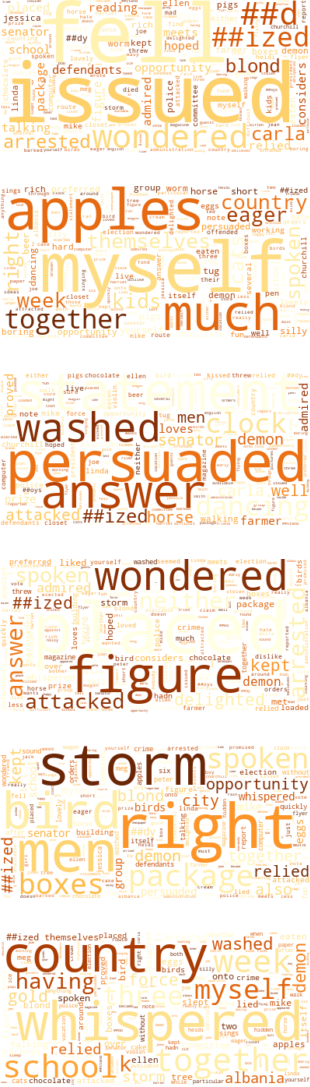}
      %\caption{model architecture of vector linear}
      \caption{Word cloud of CoLA in layer 0 to layer 6.}
      \label{wc:cola_0-6}
\end{figure}

\begin{figure}
     \centering
      \includegraphics[width=0.4\textwidth]{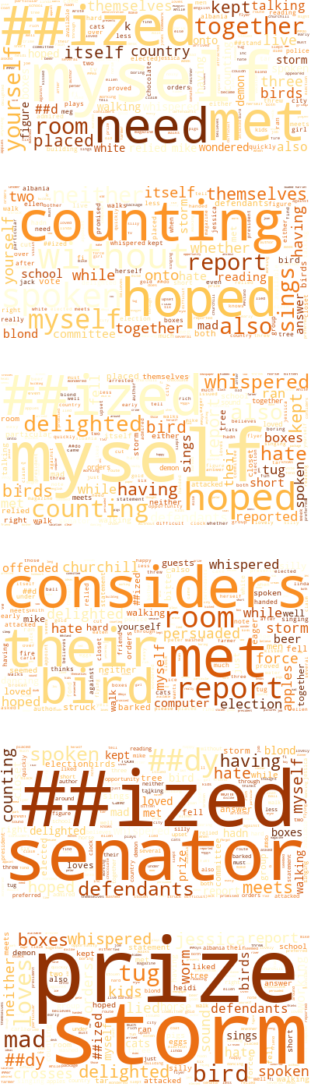}
      %\caption{model architecture of vector linear}
      \caption{Word cloud of CoLA in layer 7 to layer 12.}
      \label{wc:cola_7-12}
\end{figure}

\label{sec:appendix}

\end{document}